\documentclass{article}
\usepackage[numbers,sort&compress]{natbib}
\usepackage{spconf,amsmath,graphicx}
\usepackage[ruled,linesnumbered]{algorithm2e}
\usepackage{subfigure}

\title{ACTIVE HEADREST COMBINED WITH A DEPTH CAMERA-BASED EAR-POSITIONING SYSTEM}
\name{Yuteng Liu$^{1}$, Haowen Li$^{1}$, Haishan Zou$^{1}$, Jing Lu$^{1,2}$, Zhibin Lin$^{1}$}

\address{$^{1}$Key Laboratory of Modern Acoustics, Nanjing University, Nanjing 210093, China\\
	$^{2}$NJU-Horizon Intelligent Audio Lab, Horizon Robotics, Beijing, Beijing 100094 China\\
	\{yutengliu, haowenli\}@smail.nju.edu.cn, \{hszou, lujing, zblin\}@nju.edu.cn}

\begin{document}
\ninept
\maketitle
\begin{abstract}
Active headrests can reduce low-frequency noise around ears based on active noise control (ANC) system. Both the control system using fixed control filters and the remote microphone-based adaptive control system provide good noise reduction performance when the head is in the original position. However, their performance degrades significantly when the head is in motion. In this paper, a human ear-positioning system based on the depth camera is introduced to address this problem. The system uses RTMpose model to estimate the two-dimensional (2D) positions of the ears in the color frame, and then derives the corresponding three-dimensional (3D) coordinates in the depth frame with a depth camera. Experimental results show that the ear-positioning system can effectively track the movement of ears, and the broadband noise reduction performance of the active headrest combined with the system is significantly improved when the human head is translating or rotating.
\end{abstract}
\section{Introduction}
\label{sec:intro}

The local ANC system can generate quiet zones in noisy environments, and the quiet zone is limited in extent, with the diameter of the 10 dB quiet zone being no more than one-tenth of the wavelength of the sound wave in a diffuse sound field for a single-channel control system \cite{RN34,RN36,RN37,RN10,RN35}. The active headrest is a widely applied form of local ANC that produces a quiet zone around the human ear, which can be used in automobiles, aircraft, living rooms, and other scenarios \cite{RN36}. When the head moves, the human ears tend to deviate from the ‘quiet zone’, leading to a decrease in the effectiveness of noise reduction \cite{RN11}. Therefore, it’s necessary to track the ears' position to maintain the effective noise reduction.

Different methods have been proposed to improve the robustness of active headrests to head movements. Elliott et al. used Microsoft Kinect (Microsoft, Washington, DC, USA) to track the position of the human head and then updated the plant responses and the estimation filters in the remote microphone algorithm with the obtained position information \cite{RN12,RN13}. It is shown that the approach can achieve good noise reduction performance as the human head moves to different positions. However, they only considered the head translation and focused on the frequency range below 1 kHz. Chang et al. installed a thermal imaging module utilizing infrared technology on the airplane seat to determine the position of the passenger’s head. Nonetheless, the positioning accuracy achieved by temperature sensing is sensitive to the environment and constrained by the image resolution of the sensor \cite{RN14}. R. Han et al. designed a head positioning system with infrared range sensors, and the experimental results showed that broadband noise can be effectively controlled when the head is translated or rotated. However, the positioning system requires as many as 9 sensors to localize the head when it is translating in a horizontal plane \cite{RN17}. Behera et al. attached two microphones to a head attachment such as a hat or band, allowing both error microphones to move along with the head \cite{RN15}. In the study by T. Xiao et al., the ear of a dummy head was tracked by a camera and the sound in the ear was determined in real time by using a laser Doppler vibrometer and a lightweight reflection wafer near the cavum concha \cite{RN16}. However, a wafer or a hat needs to be worn in the above two scenarios, which is inconvenient and even impractical in practice.

In general, previous methods have encountered either limited practicality or very complex system in head-tracking.  Furthermore, the investigators predominantly focused on the translational motion of the human head, with limited attention given to the rotational motion, which is common in practice. Even head translation has been considered only in the two-dimensional (2D) plane, and the head movement in three-dimensional (3D) space has rarely been studied. 

This paper proposes an active headrest combined with a novel ear-positioning (EP) system. The EP system uses a Human pose estimation (HPE) model to identify the positions of both ears in the color frame, followed by acquiring their corresponding coordinates in the depth frame with a depth camera. To address extreme scenarios like one ear being occluded due to head rotation and involving multiple individuals in the frame, a post processing method is proposed, which estimates the 3D coordinates of the occluded ear based on the coordinates of the unoccluded ear. The active headrest system records the control filters of the human ear at different positions in the training stage, and in the control stage, the human ear position is obtained through the EP system and the corresponding control filter is used, which can maintain the robustness of the noise reduction performance when the human ears are in different positions.


\section{THE EAR-POSITIONING SYSTEM}
\label{sec:system}
\begin{figure}[!htb]
\begin{minipage}[b]{1.0\linewidth}
  \centering
  \centerline{\includegraphics[width=8.5cm]{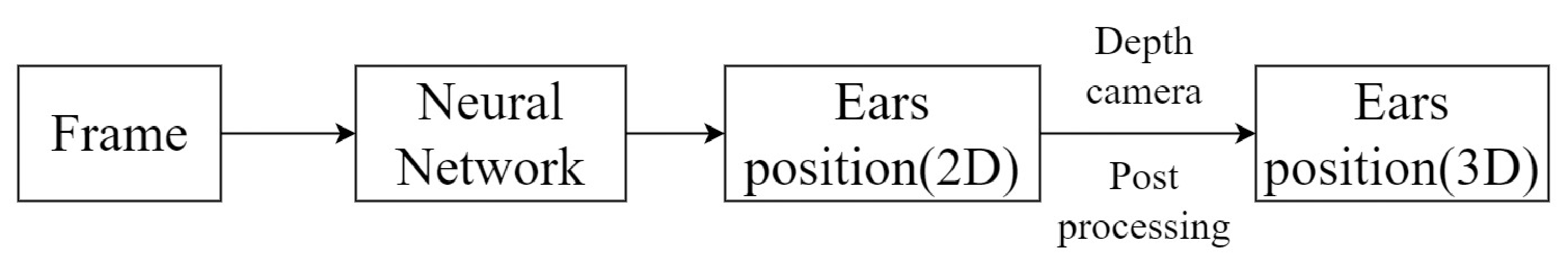}}
\end{minipage}
\caption{The pipeline of the proposed EP system}
\label{pipeline}
\end{figure}

Fig. \ref{pipeline} shows the pipeline of the proposed EP system. We employed an HPE network to perform keypoints detection on the color frame captured by the depth camera, obtaining the 2D coordinates of the ears, eyes, and nose in the frame. Subsequently, the depth frame from the depth camera is used to acquire the corresponding 3D coordinates of the detected keypoints. 
 
\subsection{Human pose estimation}
\label{ssec:HPE}
HPE is a computer vision technique that analyzes visual data to determine the spatial arrangement and localization of key body joints and parts in an image or video, including joints, limbs, and sometimes facial features. It plays a crucial role in understanding human movement, posture and interaction in various applications, such as activity recognition, motion analysis, virtual reality, augmented reality, etc \cite{RN18}. 

Modern HPE methods usually utilize deep learning algorithms to achieve high accuracy and robustness. Classic methods, such as CPN, HRNet, Vitpose \cite{RN19,RN20,RN21} have achieved impressive performance on public benchmarks. However, they suffer from bulky model sizes and high latency in practical applications \cite{RN1}. RTMpose is a high-performance real-time multi-person pose estimation framework, which achieves excellence in balancing model performance and complexity, and can be deployed on various devices (CPUs, GPUs, and mobile devices) for real-time inference \cite{RN1}. It follows the top-down paradigm, which means an off-the-shelf detector can be used to obtain bounding boxes and then estimate the pose of each person individually. Using a dedicated designed CSPNeXt as the backbone, it shows a good balance between speed and accuracy and is deployment-friendly \cite{RN23}. Furthermore, RTMPose predicts keypoints using a SimCC-based algorithm that has a very simple structure and treats keypoint localization as a classification task to reduce the quantization error \cite{RN24}. Hence, we employ RTMpose for human pose estimation to obtain the desired 2D coordinates of the ears, eyes, and nose.

These models are trained on large datasets (MS COCO \cite{RN25}, AI Challenger \cite{RN28}, MPII \cite{RN29} etc.) containing annotated human poses, allowing them to learn intricate patterns and relationships between different body parts. The types and quantity of keypoints can be defined by operators. Only five keypoints (ears, eyes and nose) need to be considered in the application scenario of this paper.

\subsection{Depth camera}
\label{ssec:DC}
After obtaining the 2D coordinates of keypoints in the color image through the HPE network, the depth frame of the depth camera is used to acquire their corresponding real-world 3D coordinates. Currently, the main solutions for depth cameras involve structured light and time-of-flight methods \cite{RN31}. Considering that active headrest application requires high positioning accuracy while not requiring long-range localization, the structured light depth camera, Intel Realsense D455 was selected for this paper \cite{RN30}. Structured light refers to the process of projecting predesigned known patterns on a scene and then capturing the images to calculate the depth for 3D surface reconstruction.

\subsection{Proposed post processing}
\label{ssec:post process}

\begin{figure}[htb]
\begin{minipage}[b]{1.0\linewidth}
  \centering
  \centerline{\includegraphics[width=8cm, height=5.5cm]{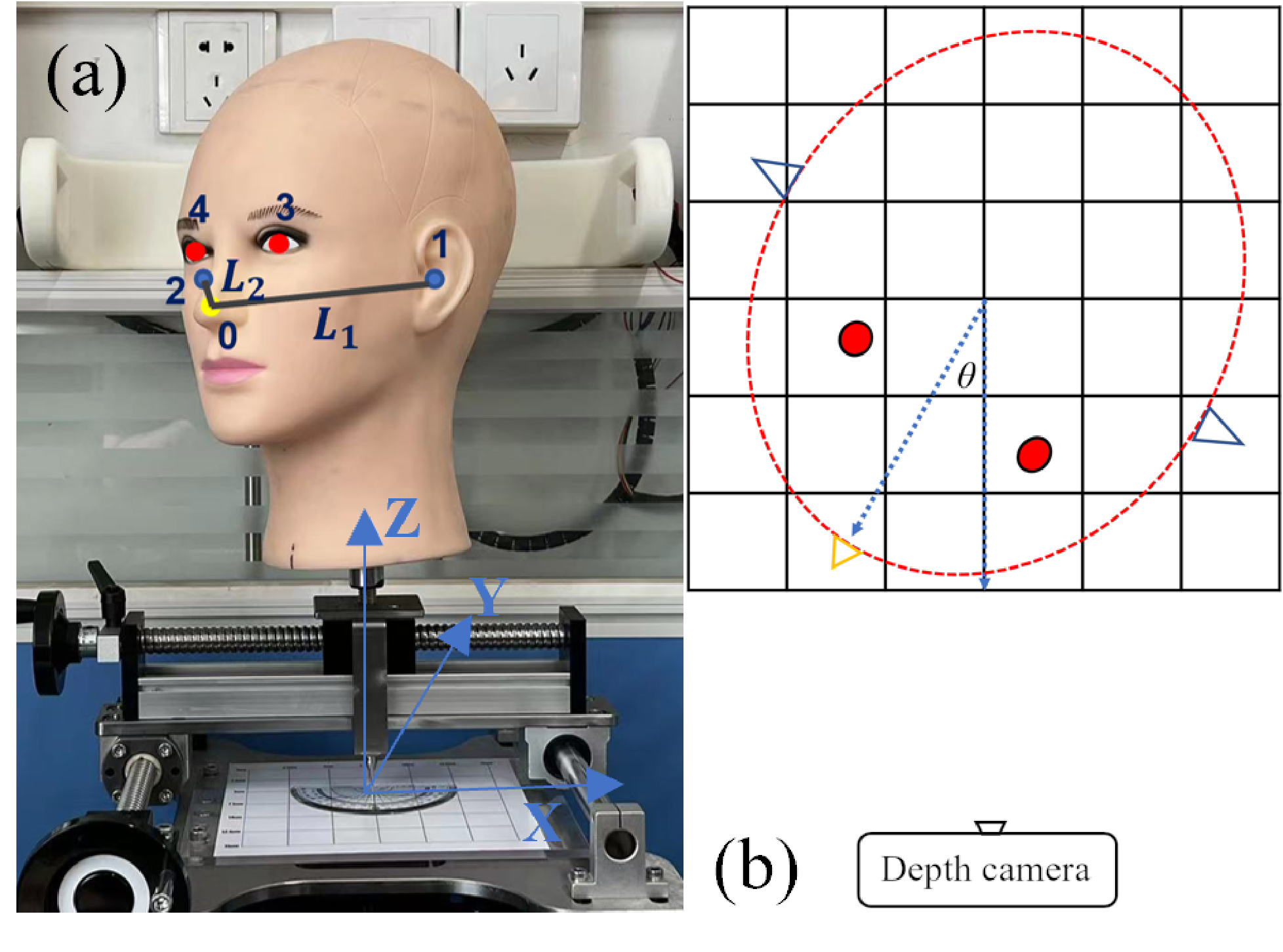}}
\end{minipage}
\caption{(a) A photo of the EP system from the front view. \\(b) A schematic diagram of the EP system from the top view.}
\label{setup}
\end{figure}

\noindent We can obtain the 3D coordinates of the target keypoints through the above procedure. However, in cases where the keypoints are located in shadowed areas that cannot be directly illuminated by the structured light, the acquired 3D coordinates may not be accurate. An example is shown in Fig. \ref{setup} (a), where only the left ear (point 1) is visible in the image due to the head rotation. The 2D HPE model can predict the position of the occluded right ear as shown by point 2. However, this point is actually placed on the edge of the face in the color image, which is not the actual right ear position.
\begin{algorithm}
\caption{Inference both ears' real 3D coordinates}\label{algorithm}
\KwData{The 3D coordinates of the five keypoints obtained from the depth camera $p_0,p_1,p_2,p_3,p_4$}
\KwResult{Both ears' real 3D coordinates $p'_1,p'_2$}
$L_1 \leftarrow |p_1-p_0|$\;
$L_2 \leftarrow |p_2-p_0|$\;
\emph{construct the perpendicular bisector plane $A$ of the $p_3$ and $p_4$}\;
\eIf{$L_1 \geq L_2$}{$p'_1 \leftarrow p_1$\;$p'_2 \leftarrow$symmetric point of $p'_1$ respect to $A$}{$p'_2 \leftarrow p_2$\;$p'_1 \leftarrow$symmetric point of $p'_2$ respect to $A$}
\end{algorithm}

To make sure the positioning system work well in any scenario, we propose a post processing algorithm based on face symmetry as shown in Algorithm 1. Firstly, we determine which ear is occluded. The Euclidean distance $L_1$ between the 3D coordinates of point 0 (the nose) and point 1 is compared with the Euclidean distance $L_2$ between point 0 and point 2. If $L_1$ exceeds $L_2$, it indicates that the right ear associated with point 2 is occluded, and vice versa, it indicates the occlusion of the left ear, corresponding to point 1. Next, the perpendicular bisector of the line connecting the points of both eyes is constructed in 3D space. Finally, the symmetric point of the unoccluded ear with respect to this perpendicular bisector and its 3D coordinates are determined, which are the actual 3D coordinates of the occluded ear.

\section{EXPERIMENTS}
\label{sec:exp}
\subsection{Positioning accuracy of the ears}
\begin{figure}[tb]
\begin{minipage}[b]{1.0\linewidth}
  \centering
  \centerline{\includegraphics[width=8.5cm]{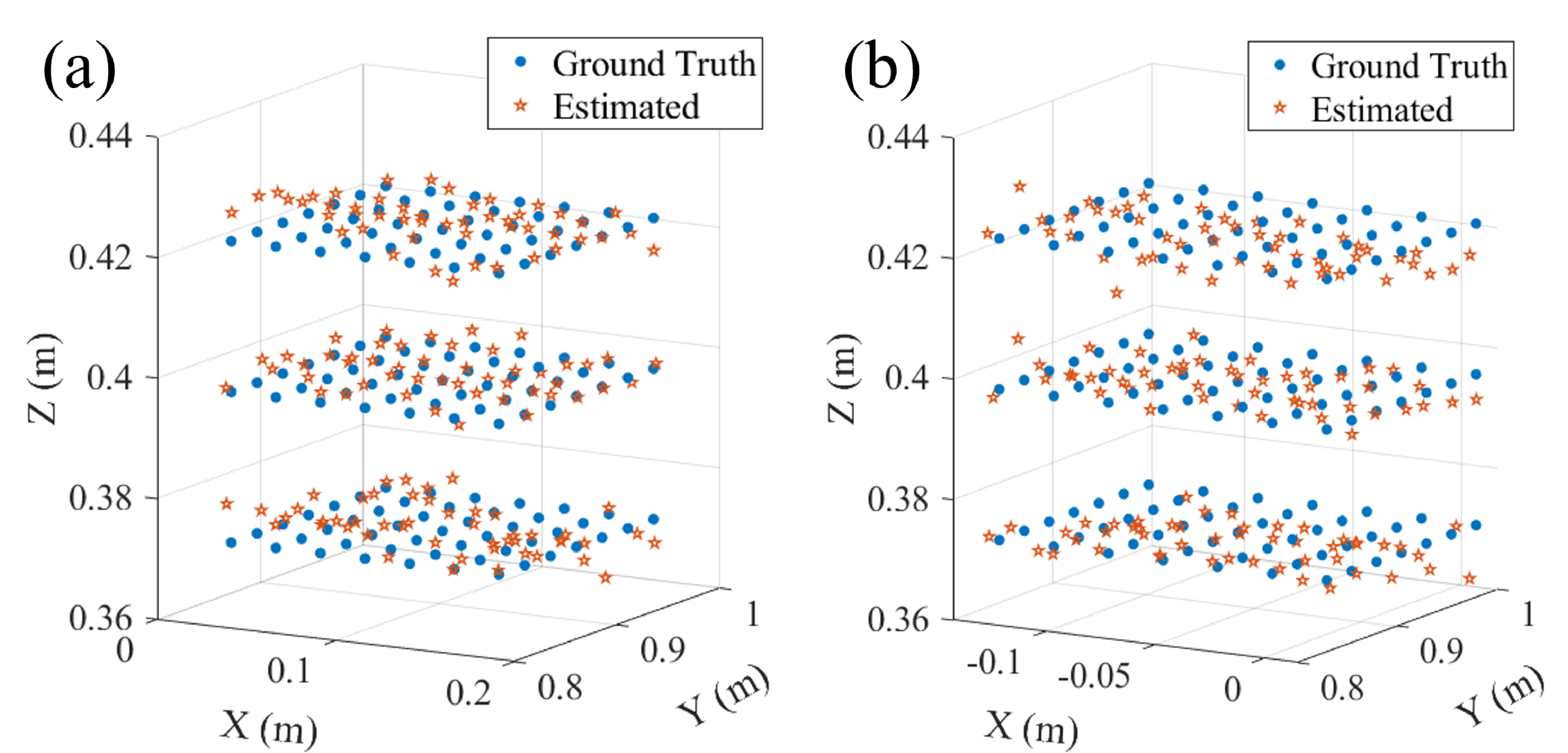}}
\end{minipage}
\caption{Comparison between the estimated positions of both ears and their real positions. (a) Left ear (b) Right ear }
\label{3Dear}
\end{figure}
 Fig. \ref{setup} shows the experimental setup for ear positioning, including a camera and a dummy head mounted on a scanning stage. The scanning stage allows the dummy head to move in a 2D plane, and a grid indicating where the dummy head resides is marked on the stage. In addition, the dummy head can be adjusted in height and rotated to different angles by means of its support shaft. Thus, the device can perform a localization experiment in 3D space. To evaluate the localization accuracy of the ear-positioning system, we conducted a grid experiment. In the experiment, the center of the dummy head was moved in translation within a $7\times7\times3$ grid, with a spacing of 2.5 cm. When the dummy head is in the initial position, the center of the dummy head coincides with the center of the grid. Fig. \ref{3Dear} presents the estimated binaural positions in comparison to their true positions, and shows that the deviations are generally less than 1 centimeter in all three dimensions. 
 As shown in Table 1, the mean deviation errors (MDE) between estimated positions and real positions of the dummy head in the XYZ directions are all less than 4 mm. 
\begin{table}[t]
\caption{Mean deviation error (MDE) for head translation.}\label{table1}
\scalebox{0.9}{\begin{tabular}{cccccccc}
\hline
\multicolumn{8}{c}{Data for movement along the X-axis. (mm)}                                                                    \\ \hline
\multicolumn{1}{c|}{Offset} & -75.0         & -50.0        & -25.0     & 0      & 25.0    & 50.0           & 75.0        \\
\multicolumn{1}{c|}{MDE (Right ear)} & 1.6         & 1.1        & 0.7     & 1.2    & 0.3   & 0.7          & 1.2       \\
\multicolumn{1}{c|}{MDE (Left ear)}  & -2.2        & -1.5       & -2.2    & -1.6   & -1.2  & -1.8         & -2.3      \\ \hline
\multicolumn{8}{c}{Data for movement along the Y-axis. (mm)}                                                                    \\ \hline
\multicolumn{1}{c|}{Offset} & -75.0         & -50.0        & -25.0     & 0      & 25.0    & 50.0           & 75.0        \\
\multicolumn{1}{c|}{MDE (Right ear)} & 4.0         & 3.1        & 3.9     & 2.0    & 4.3   & 2.9          & 2.9       \\
\multicolumn{1}{c|}{MDE (Left ear)}  & 2.1         & 1.5        & 2.4     & 0.7    & 3.9   & 1.7          & 1.8       \\ \hline
\multicolumn{8}{c}{Data for movement along the Z-axis. (mm)}                                                                    \\ \hline
\multicolumn{1}{c|}{Offset} & \multicolumn{2}{c}{-25.0}  & \multicolumn{3}{c}{0}    & \multicolumn{2}{c}{25.0}   \\
\multicolumn{1}{c|}{MDE (Right ear)} & \multicolumn{2}{c}{1.6}  & \multicolumn{3}{c}{0.2}  & \multicolumn{2}{c}{2.0}  \\
\multicolumn{1}{c|}{MDE (Left ear)}  & \multicolumn{2}{c}{-1.0} & \multicolumn{3}{c}{-1.5} & \multicolumn{2}{c}{-1.5}\\ \hline
\end{tabular}}
\end{table}

For the case of the rotation of the dummy head, the positioning error ($E_L$ and $E_R$) is defined as the 3D Euclidean distance between the estimation result and the ground truth. Since the human head is left-right symmetric, Table \ref{table2} lists the positioning error of the system when the head rotated to the right with its center fixed at the initial position. The positioning error is seen to generally increase with the rotation angle, and is less than 1 cm for rotation angles up to 45 degrees, and less than 1.4 cm for rotation angles up to 60 degrees. It is worth noting that the positioning error for the occluded right ear is not higher than that for the unoccluded left ear.

By deploying the trained model on a computer with a CPU of i5-12400@2.50GHz, the localization speed is approximately 32 frames per second (FPS), which demonstrates that the system can track the binaural positions in real time.
\begin{table}[htb] 
    \caption{Positioning error for rightward head rotation}\label{table2}
    \centering
    \scalebox{1}{\begin{tabular}{c c c}
        \hline
        $\theta(^{\circ})$ &  $E_L$(mm) & $E_R$(mm)\\ \hline
        15 &  4.0 & 5.6\\ \hline
        30 &  8.3 & 4.6\\ \hline
        45 &  9.8 & 5.9\\ \hline
        60 &  13.5 & 13.2\\   \hline
    \end{tabular}}
 
\end{table}
\vspace{-0.1cm}

\subsection{ANC headrest with ear-positioning system}
\begin{figure}[htb]
\begin{minipage}[b]{1.0\linewidth}
  \centering
  \centerline{\includegraphics[width=8.5cm, height=6cm]{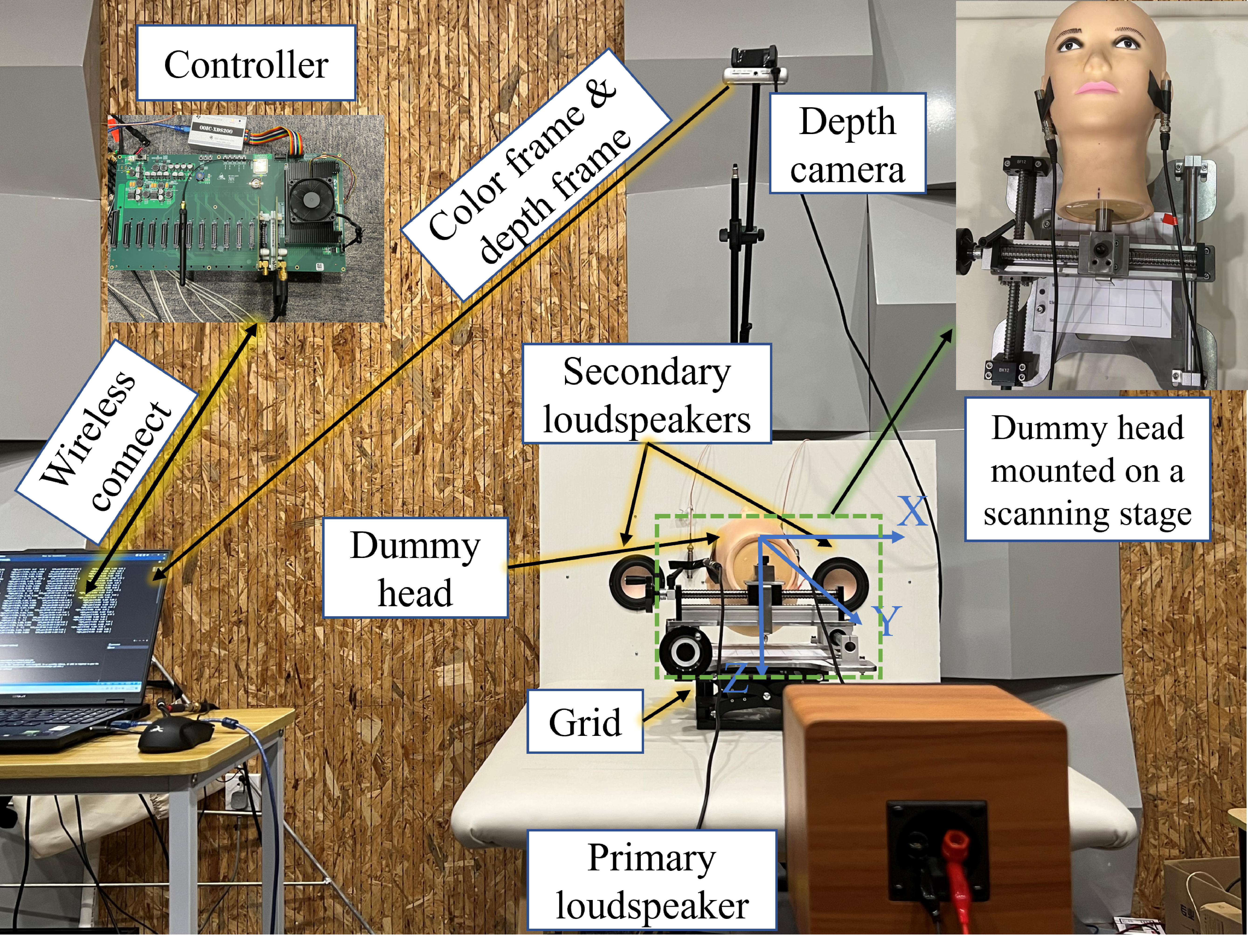}}
\end{minipage}
\caption{The experimental setup of active headrest.}
\label{anc}
\end{figure}
The experimental setup of the feedforward active headrest with the EP system is shown in Fig. \ref{anc}. The dummy head mounted on a scanning stage was placed on the bed, and a 3.5-inch loudspeaker was placed approximately 1.5 m from the center of the dummy head as the primary source, with a reference microphone (invisible in the photo) placed in its proximity. Two 3.5-inch secondary loudspeakers were placed on either side of the dummy head and two error microphones were adhered to both ears of the head. Because the active headrest uses fixed control filters for active control, error microphones were used in the training stage, while they do not operate during the control stage. The controller of the active headrest is based on a TI TMS320C6678 DSP board, and its control filters were calculated by the built-in FxLMS algorithm \cite{RN33} and fixed after convergence.

For the case of head translation, the center of the head translated in two horizontal planes, resulting in a $5\times5\times2$ grid at a 2.5 cm interval. The initial position (0,0,0) of the dummy head was at the center of the top layer (Z = 0 cm). For the case of head rotation, the dummy head was located in the center of the lower horizontal plane (Z = 2.5 cm) and rotated to the right with a maximum angle of 60 degrees. 

The following is an example of head translation to illustrate the experimental process. The ideal control filters were trained and stored in the controller when the dummy head translated to each grid point in the training stage. In the control stage, when the dummy head was moved, for the case where the EP system was off, the control filter obtained at the initial position was used. For the case where the EP system was on, the coordinates obtained by the EP system were transmitted to the controller through a wireless protocol interface. The controller determined the closest grid point to that coordinates and selected the pre-stored control filter associated with it for control.
\begin{figure}[htb]
\begin{minipage}[b]{1.0\linewidth}
  \centering
  \centerline{\includegraphics[width=8.5cm]{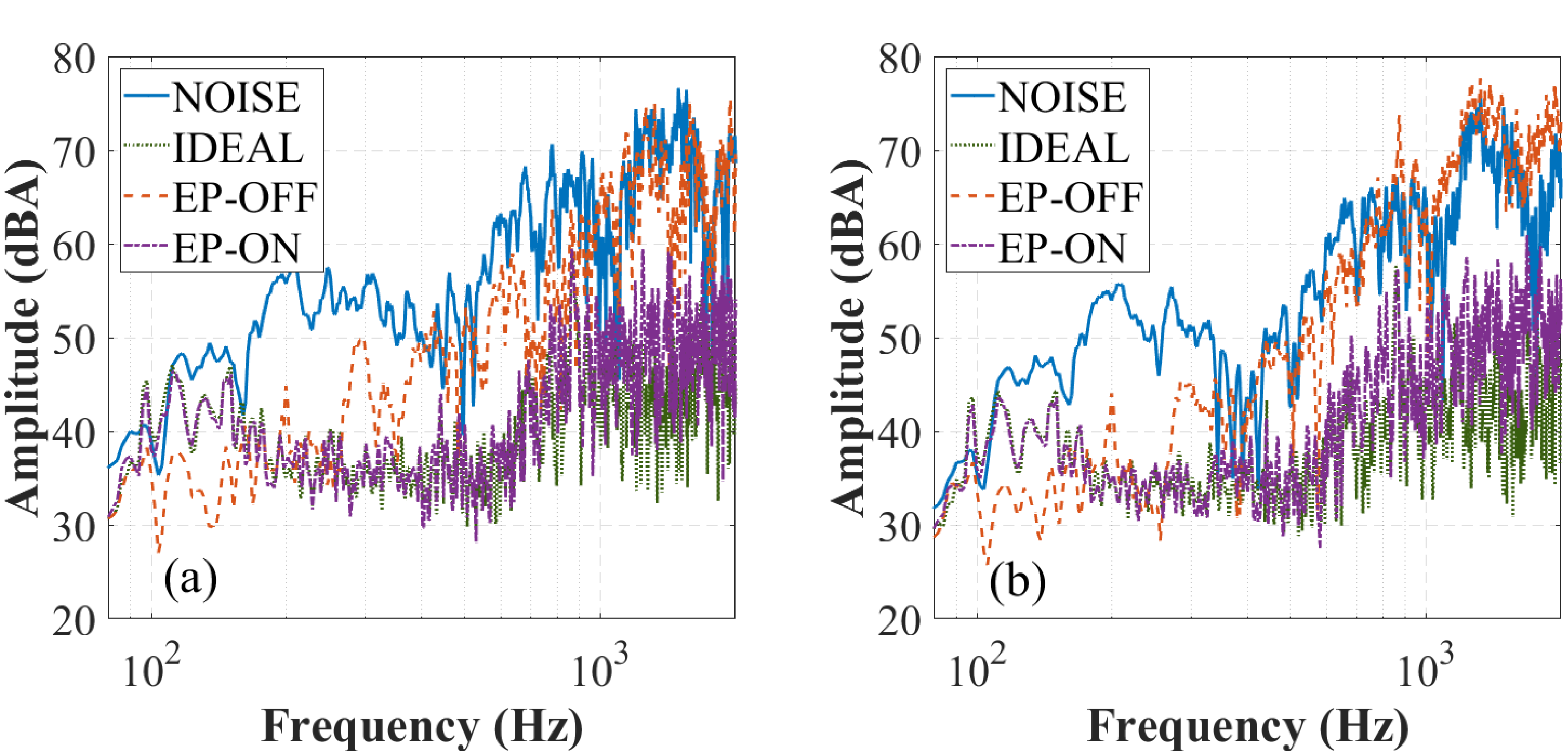}}
\end{minipage}
\caption{The sound pressure level spectra at both ears with the center of the dummy head at (-2.5, -2.5, 2.5) cm. (a) Left ear (b) Right ear}
\label{fre}
\end{figure}

At each grid point, the noise reduction was measured under three conditions: using the ideal control filter with microphones attached to the ears acting as the error sensors, using the control filter with the EP system off, and using the selected control filter with the EP system on. Fig. \ref{fre} shows the sound pressure level spectra before and after noise control when the dummy head was at (-2.5, -2.5, 2.5) cm, where the frequency range was 80-2000 Hz. It can be seen that with the EP system off, the head movement results in a considerable decrease in noise reduction, especially at high frequencies. The noise reduction performance was improved with the EP system on, approaching that of the ideal filter in that position.

\begin{figure}[!htb]
\begin{minipage}[b]{1.0\linewidth}
  \centering
  \centerline{\includegraphics[width=8.5cm]{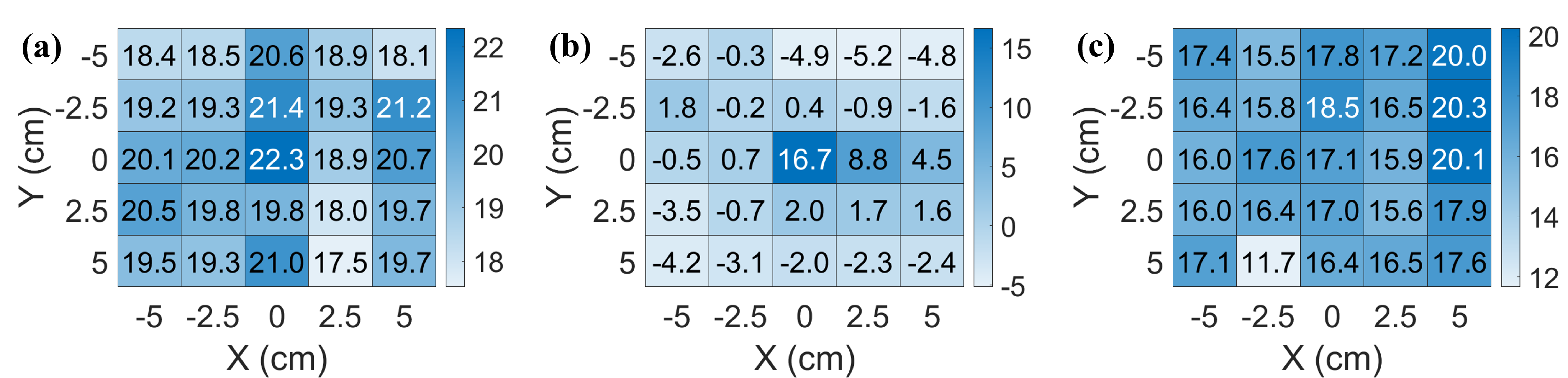}}
\end{minipage}
\subfigure{\begin{minipage}[b]{1.0\linewidth}
  \centering
  \centerline{\includegraphics[width=8.5cm]{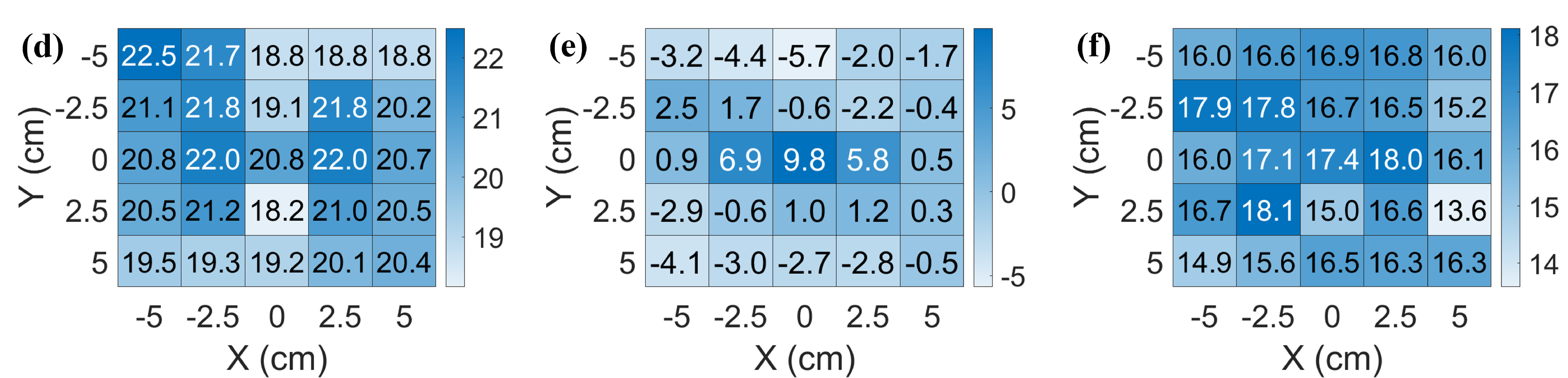}}
\end{minipage}}
\caption{The noise reduction (dBA) of left ear at different positions and in different control situations. (a-c) Layer 1 (d-f) Layer 2 (a,d) Ideal (b,e) EP-OFF (c,f) EP-ON}
\label{anc_result_left}
\end{figure}

\begin{figure}[!htb]
\begin{minipage}[b]{1.0\linewidth}
  \centering
  \centerline{\includegraphics[width=8.5cm]{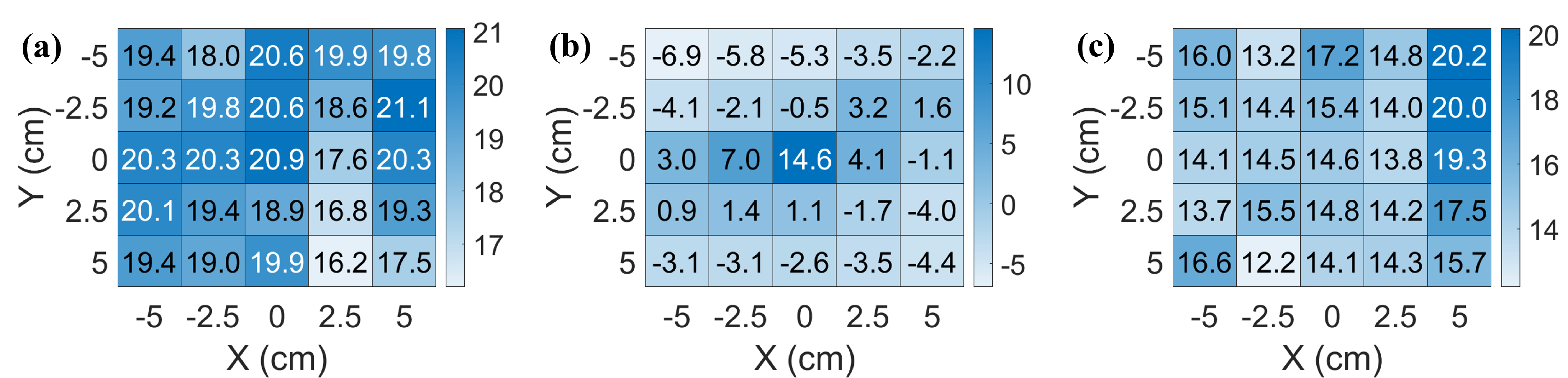}}
\end{minipage}
\subfigure{\begin{minipage}[b]{1.0\linewidth}
  \centering
  \centerline{\includegraphics[width=8.5cm]{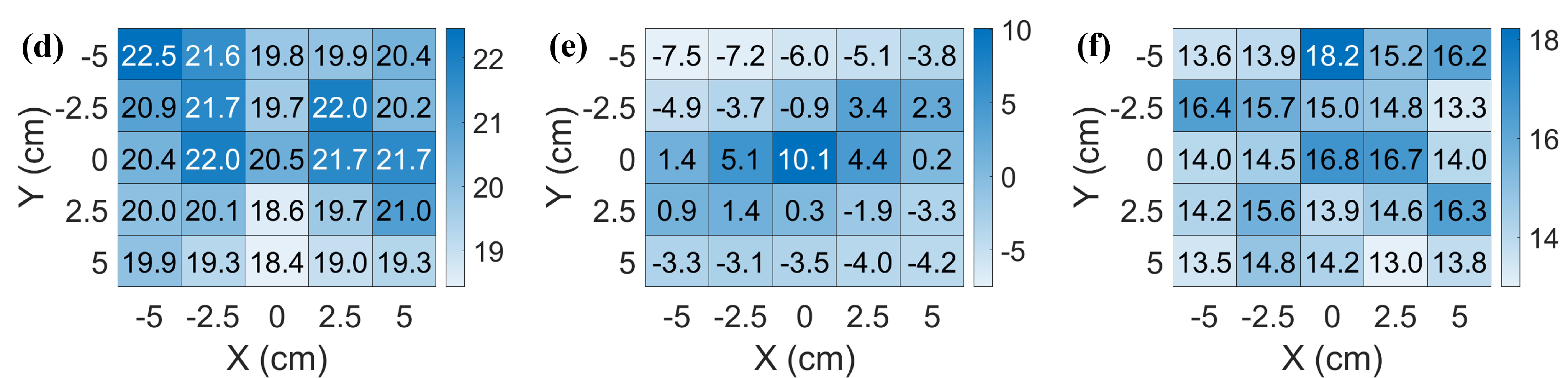}}
\end{minipage}}
\caption{The noise reduction (dBA) of right ear at different positions and in different control situations. (a-c) Layer 1 (d-f) Layer 2 (a,d) Ideal (b,e) EP-OFF (c,f) EP-ON}
\label{anc_result_right}
\end{figure}

Fig. \ref{anc_result_left} and Fig. \ref{anc_result_right} show the noise reduction (NR) measured at the left and right ear, respectively, during the dummy head translation, where the numbers in squares are in dBA. Figures (a-c) show the results for layer 1 (Z = 0 cm) and Figures (d-f) for layer 2 
\vspace{-0.3cm}
(Z = 2.5 cm). The NR in Figs. (a) and (d) represent the utilization of the ideal control filter at each measurement point. The NR in Figs. (b) and (e) correspond to the use of the control filter pre-trained at the initial position of the dummy head. It decreased significantly as the head deviated from the initial position, with a maximum variance of 22.4 dBA compared to the NR at the initial position. Figs. (c) and (f) show the NR in the case where the proposed EP system was turned on and the control filter was employed based on the ear position obtained by the system. It can be seen that the NR significantly improved compared to that obtained without EP system. 
\begin{figure}[tb]
\vspace{0.1cm}
\begin{minipage}[b]{1.0\linewidth}
  \centering
  \centerline{\includegraphics[width=8.5cm]{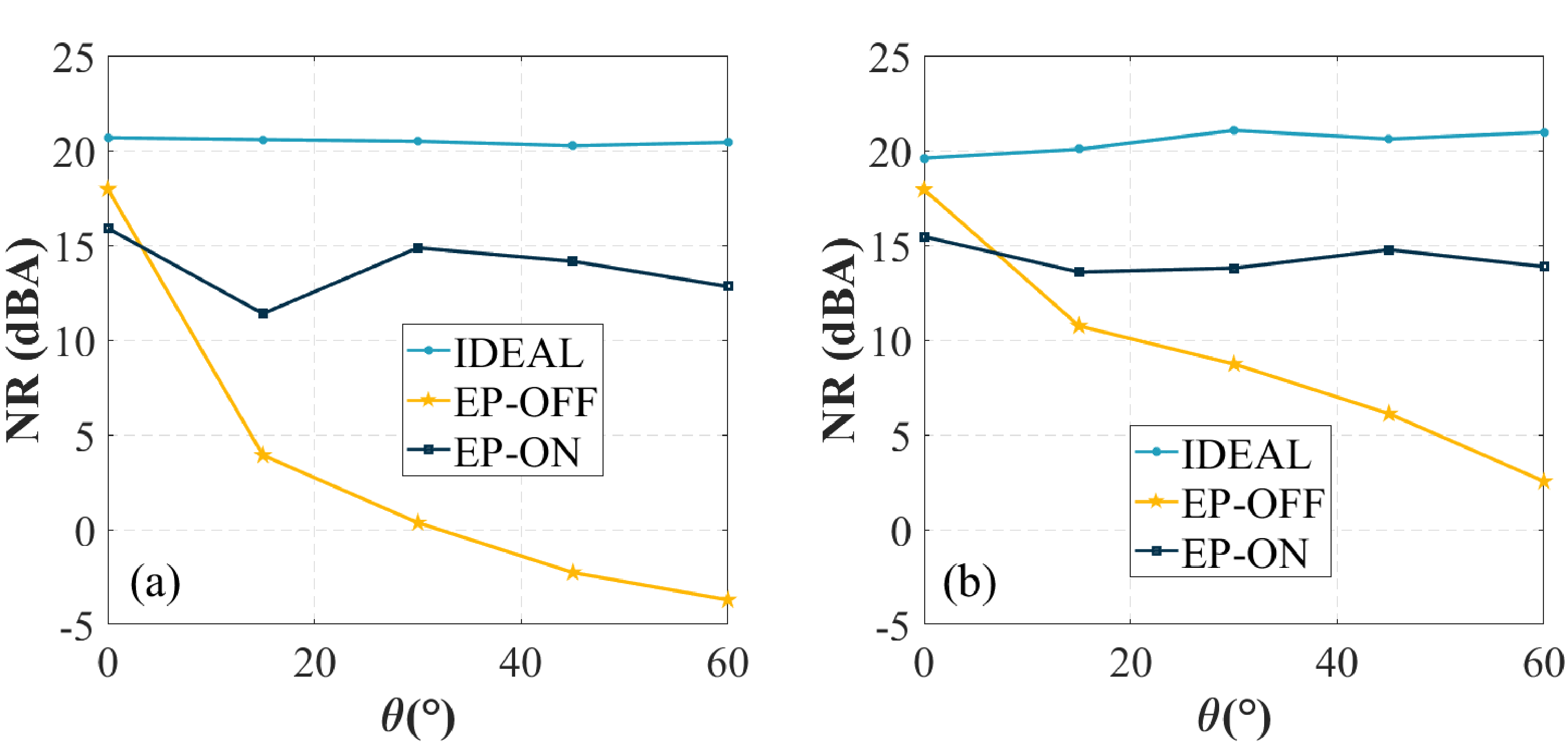}}
\end{minipage}
\caption{The noise reduction at different rotation angles. (a) Left ear (b) Right ear}
\label{rotation}
\end{figure}

Fig. \ref{rotation} displays the NR at both ears as the head rotated to different angles to the right. It is evident that, when the EP system was turned off, as the rotation angle of the head increased, the NR significantly decreased. At a rotation angle of 60 degrees, the maximum variance compared to that using the ideal control filter exceeds 23 dBA. On the other hand, when the EP system was turned on, the system's performance showed remarkable robustness, with a maximum reduction of no more than 8 dBA compared to the ideal control. The results further validate the effectiveness of the EP system.  Besides the performance of the EP system, the variance in NR between the “Ideal” and “EP-on” scenarios could potentially be attributed to the manual adjustments of the dummy head, leading to slight discrepancies in its positioning during the two movements.

\section{CONCLUSIONS}

An ear-positioning system for active headrest was designed using a depth camera and the human pose estimation model RTMpose. To address the extreme scenarios where the human ear is occluded, a post processing algorithm was proposed. The location experiments show that the mean positioning error is less than 4 mm when the head is in translational motion and less than 1.4 cm when the head rotates within the range of $\pm 60^{\circ}$. For frequency bands below 2 kHz, the active headrest based on the EP system maintains good noise reduction during head movement. The lowest noise reduction was 11.7 / 12.2 dBA with the EP system and -4.8 / -7.5 dBA without the EP system for head translations up to approximately 7.5 cm. For head rotations within the range of $\pm 60^{\circ}$, the lowest noise reduction with the positioning system was 11.4 / 13.6 dBA, much higher than the -3.7 / 2.5 dBA without the system.

\section{ACKNOWLEDGMENT}
\label{sec:ACKNOW}
This work was supported by the National Natural Science Foundation of China (Grant No. 12274221).
\clearpage
\bibliographystyle{IEEEbib}
\bibliography{strings,refs}

\end{document}